\documentclass{article}
\usepackage{graphicx} 
\usepackage{float}
\usepackage{subcaption}
\usepackage{natbib}
\usepackage{natbib}
\usepackage{amssymb}
\usepackage{amsmath}

\usepackage[a4paper, margin=1in]{geometry} % Adjust margins as needed

\title{BERT4Traj: Transformer-Based Trajectory Reconstruction for Sparse Mobility Data}
\date{} 

\author{
  Hao Yang \\
  University of Georgia \\
  Athens, GA, USA \\
  \texttt{hy96161@uga.edu}
  \and
  Angela Yao\thanks{Corresponding author: \texttt{xyao@uga.edu}} \\
  Department of Geography \\
  University of Georgia \\
  Athens, GA, USA
  \and
  Christopher C. Whalen \\
  College of Public Health \\
  University of Georgia \\
  Athens, GA, USA \\
  \texttt{ccwhalen@uga.edu}
  \and
  Gengchen Mai \\
  University of Texas at Austin \\
  Austin, TX, USA \\
  \texttt{gengchen.mai@austin.utexas.edu}
}

\begin{document}

\maketitle

\section*{Abstract}
Understanding human mobility is essential for applications in public health, transportation, and urban planning. However, mobility data often suffers from sparsity due to limitations in data collection methods, such as infrequent GPS sampling or call detail record (CDR) data that only capture locations during communication events. To address this challenge, we propose BERT4Traj, a transformer-based model that reconstructs complete mobility trajectories by predicting hidden visits in sparse movement sequences. Inspired by BERT’s masked language modeling objective and self-attention mechanisms, BERT4Traj leverages spatial embeddings, temporal embeddings, and contextual background features such as demographics and anchor points. We evaluate BERT4Traj on real-world CDR and GPS datasets collected in Kampala, Uganda, demonstrating that our approach significantly outperforms traditional models such as Markov Chains, KNN, RNNs, and LSTMs. Our results show that BERT4Traj effectively reconstructs detailed and continuous mobility trajectories, enhancing insights into human movement patterns. 

\noindent \textbf{Keywords:} Human Mobility, Trajectory Reconstruction, Deep Learning, CDR, GPS

\section{Introduction}
Understanding human mobility is crucial for various applications, including public health, transportation, and urban planning \citep{meloni2011modeling, belik2011natural}. With the increasing availability of location data from GPS devices, mobile phones, and other portable technologies, human mobility analysis has gained significant attention. However, despite the abundance of location data, data sparsity remains a persistent challenge. For example, Call Detail Records (CDRs) capture locations only when calls or text messages occur, leaving significant gaps in a user's movement trajectory \citep{chen2019complete}. Similarly, GPS data may be sparse due to battery-saving modes, signal loss, or intermittent sampling. Consequently, there exist places that individuals have visited but are not recorded in the data, which we refer to as ``hidden visits'' \citep{barnett2020inferring}. The presence of hidden visits impedes the ability to reconstruct a complete view of an individual's daily movement, posing substantial challenges to understanding human mobility. Thus, identifying hidden visits to address data sparsity and reconstructing continuous, detailed, and complete mobility trajectories is a necessary and meaningful research problem.

Early trajectory reconstruction methods primarily relied on Markov Chains and interpolation-based techniques \citep{gambs2012next, yu2018using, huang2015predicting, hoteit2014estimating}. Markov Chain models, such as the one proposed by \citet{huang2015predicting}, incorporate activity changes to enhance mobility prediction but struggle with long-range dependencies and complex movement behaviors due to the adopted Markov assumption. Interpolation-based approaches leverage spatial-temporal correlations to estimate missing points. For instance, \citet{hoteit2014estimating} use linear and cubic interpolation to reconstruct human mobility from mobile phone data. However, such assumptions fail to capture real-world non-linear travel patterns effectively.

With advancements in machine learning, researchers have increasingly adopted deep learning models for trajectory reconstruction \citep{chen2014traces, li2019reconstruction, liu2018mapping, wang2019deep}. Backpropagation (BP) neural networks, as proposed by \citet{liu2018mapping}, reconstruct mobility trajectories from sparse Call Detail Records (CDR) to estimate hourly population density. However, this method assumes predictable movement patterns, overlooking detours and irregular trajectories.

More recently, Transformer-based approaches have demonstrated superior performance \citep{si2023trajbert, crivellari2022tracebert}. TrajBERT, introduced by \citet{si2023trajbert}, applies BERT-based trajectory recovery with spatial-temporal refinement to address implicit trajectory sparsity. While effective in predicting missing locations, TrajBERT lacks external context modeling, such as user characteristics, dynamic temporal variations, or real-world events, limiting its adaptability for high-accuracy trajectory prediction.

To overcome these challenges, this paper introduces \textbf{BERT4Traj}, a novel Transformer-based model for trajectory reconstruction. By leveraging BERT’s bidirectional self-attention mechanism, BERT4Traj effectively captures spatial-temporal dependencies, improving trajectory prediction accuracy. The model is applied to reconstruct complete movement trajectories from both CDR and GPS datasets, demonstrating its robustness in handling data sparsity across different mobility data types. Through context-aware trajectory reconstruction, BERT4Traj enables a more detailed and accurate representation of human mobility patterns, offering valuable insights for applications in public health, urban planning, and transportation analytics.

\section{Methodology}

To address the challenge of data sparsity and reconstruct continuous, detailed mobility trajectories, we propose a transformer-based architecture, BERT4Traj, inspired by BERT. This model predicts hidden visits in user mobility trajectories by treating each user’s daily trajectory as a sequence analogous to a sentence in Natural Language Processing (NLP), where locations correspond to ordered words. The objective is to infer missing locations within this sequence using spatial, temporal, and user-specific demographic information.

The core idea of BERT4Traj is inspired by masked language modeling in BERT, where predictions are made based on contextual information from surrounding tokens. In the context of human mobility, locations visited on the same day provide contextual clues to infer missing visits. In addition to known locations, background information such as demographic characteristics (e.g., age, gender), key life anchors (e.g., home and workplace), and temporal context (e.g., weekday vs. weekend, holidays) further enrich the representation of an individual’s mobility behavior.

As illustrated in Figure 1, BERT4Traj incorporates a BERT-like masking and prediction mechanism. A subset of locations in a trajectory sequence is randomly masked, and the model learns to predict these missing locations using the context provided by the rest of the sequence. This bidirectional prediction process enables BERT4Traj to develop a deep understanding of how visited locations relate to one another in varying contexts. 

The input sequence consists not only of the trajectory data but also of unmasked context tokens that provide additional background information, including temporal attributes, user demographics, and travel characteristics. During training, the model learns intricate relationships between visited locations and contextual features, allowing it to accurately predict missing locations at specific times in a day. Ultimately, this approach reconstructs an individual’s movement trajectory with finer temporal granularity, effectively addressing data sparsity issues in mobility datasets.

\begin{figure}[htbp]
    \centering
    \includegraphics[width=1\textwidth]{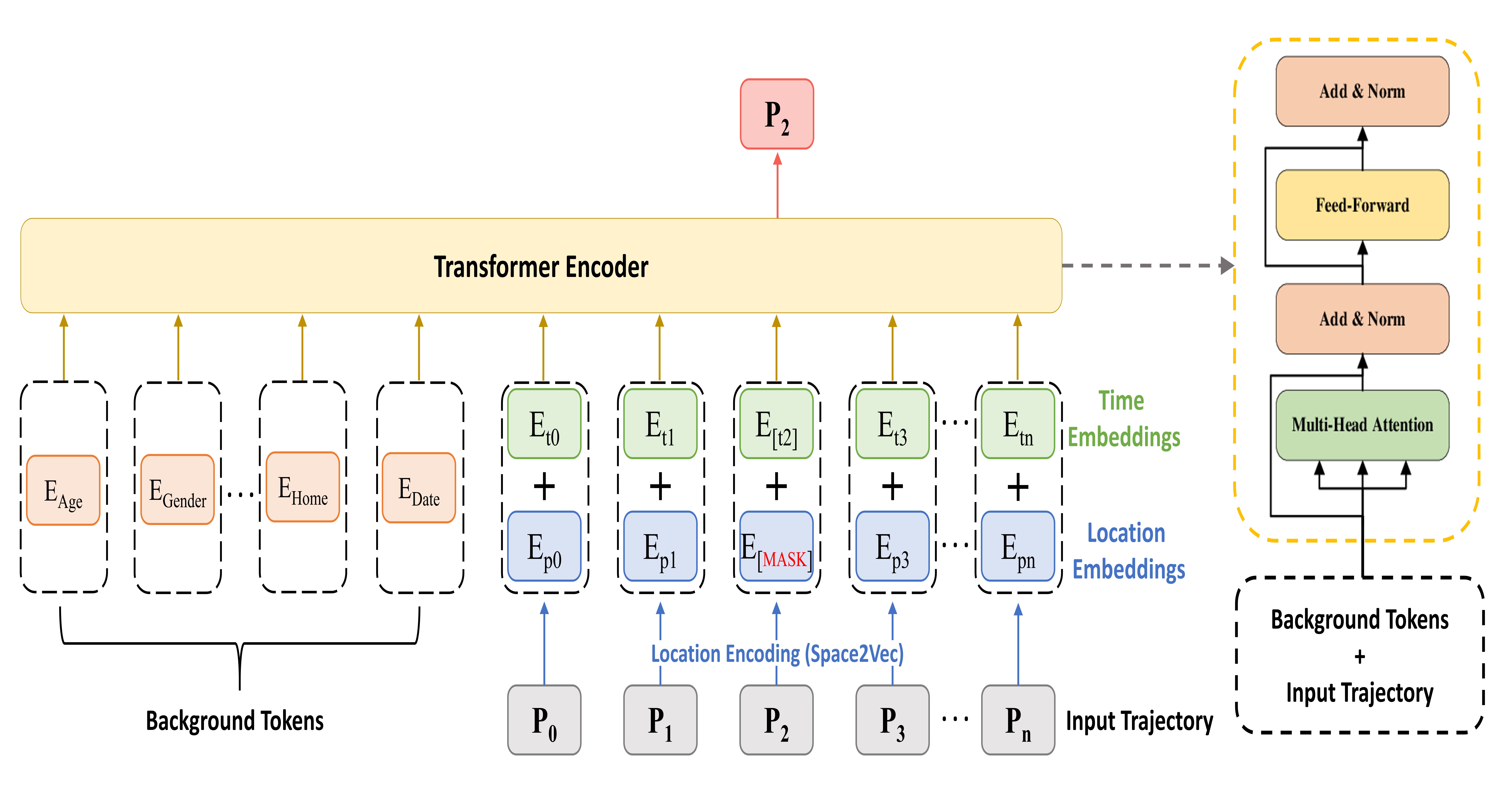} % Adjust width as needed
    \caption{The overall framework of the BERT4Traj model
}
    \label{fig:example}
\end{figure}

\subsection{Data Representation and Embeddings}
Each user's daily trajectory consists of a sequence of visited locations with corresponding timestamps. To represent this information in the model, we define location embeddings and time embeddings. Each visited location is embedded as a vector of dimension \( d \):
\begin{equation}
    \mathbf{l}_i \in \mathbb{R}^{d},
\end{equation}
where \( i \) is the index of the location in the trajectory. The location embeddings encode geographical information (latitude, longitude) and may also include semantic attributes, such as Points of Interest (POI) types or travel modes, if available.

To model temporal dependencies, a time embedding \( \mathbf{t}_i \in \mathbb{R}^{d} \) is generated based on the timestamp capturing the time of the visit. The time embedding functions similarly to positional embeddings in NLP models, providing temporal context to the trajectory sequence.

In addition to trajectory embeddings, we incorporate background tokens representing demographic features, anchor points, and temporal attributes. Specifically:
\begin{equation}
    \mathbf{B} = [\mathbf{w}_{\text{age}}; \mathbf{w}_{\text{gender}}; \dots]
\end{equation}
represents the demographic embeddings, where \( \mathbf{w}_{\text{age}} \) and \( \mathbf{w}_{\text{gender}} \) denote the age and gender embeddings, respectively.

Anchor points, such as home and workplace locations, are encoded as:
\begin{equation}
    \mathbf{A} = [\mathbf{w}_{\text{primary}}; \mathbf{w}_{\text{secondary}}; \dots]
\end{equation}
where \( \mathbf{w}_{\text{primary}} \) and \( \mathbf{w}_{\text{secondary}} \) represent embeddings for primary and secondary anchor points.

Temporal context, including whether the day is a weekday, weekend, or holiday, is represented as:
\begin{equation}
    \mathbf{T} = [\mathbf{w}_{\text{weekday}}; \mathbf{w}_{\text{weekend}}; \dots]
\end{equation}
where \( \mathbf{w}_{\text{weekday}} \) and \( \mathbf{w}_{\text{weekend}} \) denote the corresponding time-related embeddings.

The complete input sequence is formulated as:
\begin{equation}
    \mathbf{X} = [\mathbf{B}; \mathbf{A}; \mathbf{T}; \mathbf{l}_1 + \mathbf{t}_1; \mathbf{l}_2 + \mathbf{t}_2; \dots; \mathbf{l}_n + \mathbf{t}_n].
\end{equation}

\subsection{Masking Mechanism}
A portion of the location tokens in the trajectory sequence is randomly masked. The objective is to predict these masked locations using unmasked locations and contextual embeddings. Let \( \mathbf{M} \in \{0,1\}^{n} \) be a binary masking vector, where \( \mathbf{M}_i = 1 \) if the location \( \mathbf{l}_i \) is masked and \( \mathbf{M}_i = 0 \) otherwise. The masked sequence is represented as:
\begin{equation}
    \mathbf{X}_{\text{masked}} = [\mathbf{B}; \mathbf{A}; \mathbf{T}; (\mathbf{M}_1 \cdot \mathbf{l}_1) + \mathbf{t}_1; \dots; (\mathbf{M}_n \cdot \mathbf{l}_n) + \mathbf{t}_n].
\end{equation}
This ensures that spatial and temporal relationships are preserved while training the model to infer missing locations.

\subsection{Transformer-Based Sequence Encoder}
The masked sequence is processed by a Transformer encoder consisting of multiple self-attention layers. The self-attention mechanism computes dependencies between different locations in the trajectory:
\begin{equation}
    \text{Attention}(Q, K, V) = \text{softmax}\left(\frac{QK^T}{\sqrt{d_k}}\right) V,
\end{equation}
where \( Q, K, V \) represent the query, key, and value matrices derived from the input sequence, and \( d_k \) is the dimensionality of the key vectors.

Multi-head attention further enhances the model's ability to capture diverse mobility patterns:
\begin{equation}
    \text{MultiHead}(Q, K, V) = \text{Concat}(\text{head}_1, \dots, \text{head}_h) W^O.
\end{equation}
The output of the encoder is a sequence of hidden states \( H = [h_1, h_2, \dots, h_n] \), where each \( h_i \) encodes contextual information about its corresponding location.

\subsection{Masked Location Prediction}
For each masked location \( \mathbf{l}_i \), the model predicts its most probable value using the output hidden states:
\begin{equation}
    \hat{\mathbf{l}}_i = \text{softmax}(W h_i),
\end{equation}
where \( W \in \mathbb{R}^{|P| \times d} \) is a weight matrix, and \( \hat{\mathbf{l}}_i \) represents the predicted probability distribution over the possible locations \( P \).

The model is optimized using a cross-entropy loss function:
\begin{equation}
    \mathcal{L} = - \sum_{i \in \mathcal{M}} \log \hat{\mathbf{l}}_i[\mathbf{l}_i].
\end{equation}
Minimizing this loss encourages the model to correctly predict masked locations, improving its ability to reconstruct missing trajectory data.

\section{Application of BERT4Traj to CDR and GPS Data}

The BERT4Traj model is applied to reconstruct mobility trajectories from two types of location data: GPS and CDR. To assess its effectiveness, we evaluate BERT4Traj on datasets collected from users in Kampala, Uganda. The CDR dataset consists of 248 participants, while the GPS dataset includes 586 participants.

\subsection{CDR Data}
CDR data captures the tower location and timestamp when a communication event occurs, such as a call or text message. Due to its event-driven nature, CDR data is sparse, providing only a limited view of daily mobility patterns. On average, each individual in our dataset has location records for only five hourly time slots per day, leaving significant gaps in their trajectory. To address this, BERT4Traj predicts hidden visits during unrecorded periods, enhancing trajectory completeness.

Each input trajectory is represented as a sequence of tower locations with timestamps. We generate location embeddings using Space2Vec \citep{mai2020multi}, which provides continuous vector representations based on geographical coordinates:
\begin{equation}
    e_{loc}(l_i) = \text{space2vec}(\mathbf{L}(p_i)),
\end{equation}
where \( \mathbf{L}(p_i) \) denotes the latitude and longitude of tower location \( p_i \). 

For time embeddings, we divide the 17-hour time window (from 6:00 AM to 11:00 PM) into 34 half-hour slots, assigning an index from 1 to 34 to each slot. We apply sinusoidal positional encoding to preserve temporal relationships:
\begin{equation}
    t_i = 
    \left\{
    \begin{array}{ll}
        \sin \left( \frac{s_i}{10000^{\frac{2j}{d}}} \right), & \quad \text{if } \, j \, \text{ is \, even} \\
        \cos \left( \frac{s_i}{10000^{\frac{2j}{d}}} \right), & \quad \text{if } \, j \, \text{ is \, odd}
    \end{array}
    \right.
\end{equation}

where:
\( s_i \) is the time slot index (ranging from 1 to 34),
\( j \) is the embedding dimension index,
\( d \) is the total embedding dimension.

This encoding ensures that nearby time slots have similar embeddings, allowing the model to recognize the temporal structure of the sequence effectively.

Additional context, such as age, gender, primary and secondary anchor points, and temporal indicators (weekday vs. weekend), is incorporated into the model. BERT4Traj predicts missing locations within these time intervals, reconstructing a temporally detailed trajectory.

\subsection{GPS Data}
GPS data has a higher temporal resolution than CDR but still contains missing records due to device limitations, such as power-saving modes, signal loss, or shutdowns. To create a continuous trajectory, we apply BERT4Traj to predict missing locations.

The study area is divided into a 100m \( \times \) 100m grid, with each cell representing a possible location. Location embeddings are derived using Space2Vec to maintain spatial coherence.

For time encoding, we use normalized time encoding to preserve the full timestamp (hour, minute) in a continuous and periodic manner. First, we normalize the time to a fraction of the day:
\begin{equation}
t_{\text{norm}} = \frac{\text{hour} \times 60 + \text{minute}}{1440}
\end{equation}
where 1440 is the total number of minutes in a day.

Then, we apply sinusoidal encoding to capture periodicity:
\begin{equation}
    t_i = 
    \left\{
    \begin{array}{c}
        \sin(2\pi t_{\text{norm}}) \\
        \cos(2\pi t_{\text{norm}})
    \end{array}
    \right.
\end{equation}

This encoding ensures that time is represented in a continuous way, maintaining smooth transitions between consecutive timestamps while preserving cyclic properties.

Similarly, we incorporate background tokens representing demographic attributes, anchor points, and temporal indicators. BERT4Traj reconstructs continuous trajectories by predicting the most likely location (grid ID) at any given time.

\subsection{Model Evaluation}
The effectiveness of BERT4Traj was evaluated against several baseline models, including Markov Chain, RNN, LSTM, and KNN, using both CDR and GPS datasets. Performance is assessed using the following metrics:
\begin{itemize}
    \item \textbf{Accuracy}: The proportion of correctly predicted locations.
    \item \textbf{Top-3 Accuracy}: Whether the correct location is among the top three predictions.
    \item \textbf{Top-5 Accuracy}: Whether the correct location is among the top five predictions.
\end{itemize}

Table 1 summarizes the performance comparison.The results clearly demonstrate that BERT4Traj outperforms all baseline models across both datasets. In the CDR dataset, BERT4Traj achieves an accuracy of 87.1\%, significantly surpassing LSTM (74.5\%) and RNN (70.6\%), highlighting its superior ability to handle sparse mobility data compared to recurrent models. Similarly, in the GPS dataset, BERT4Traj attains 71.4\% accuracy, outperforming LSTM (62.1\%) and RNN (60.3\%). The lower accuracy observed in the GPS dataset compared to CDR is due to the fundamental difference in prediction tasks—CDR reconstruction predicts locations within predefined half-hour time slots, whereas GPS trajectory reconstruction requires continuous predictions across time, making the task inherently more challenging. Despite this, BERT4Traj still achieves notable improvements over baseline models, demonstrating its robustness in handling missing data.

Among the baseline models, Markov Chain and KNN exhibit the lowest accuracy, particularly in the GPS dataset, where their accuracy remains below 55\%. This indicates that these simpler models struggle to capture sequential dependencies and complex spatial-temporal relationships, reinforcing the advantage of deep learning approaches in trajectory reconstruction.

\begin{table}[H]
    \centering
    \caption{Comparison with baselines in Accuracy, Top-3 Accuracy, and Top-5 Accuracy}
    \includegraphics[width=0.8\textwidth]{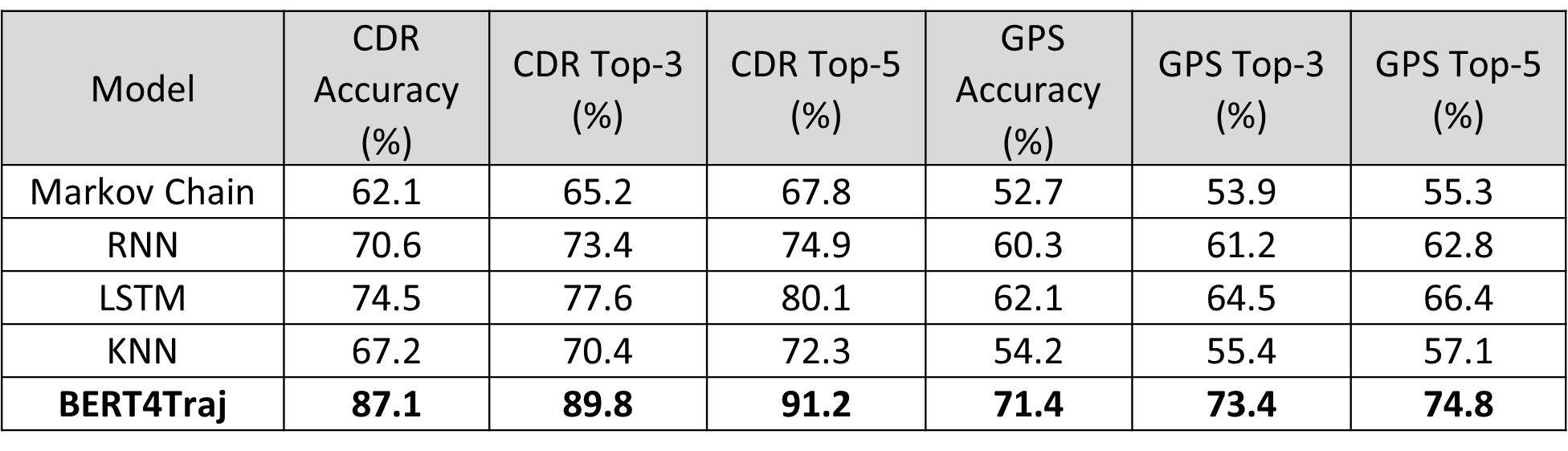}
    \label{tab:comparison_baselines}
\end{table}

To examine the contribution of different contextual background features, we conducted an ablation study where demographic information, anchor points, and date information were individually removed from BERT4Traj. The results of this analysis are shown in Table 2 below.

\begin{table}[H]
    \centering
    \caption{Ablation Study: Effect of Removing Contextual Features on Model Accuracy}
    \includegraphics[width=0.8\textwidth]{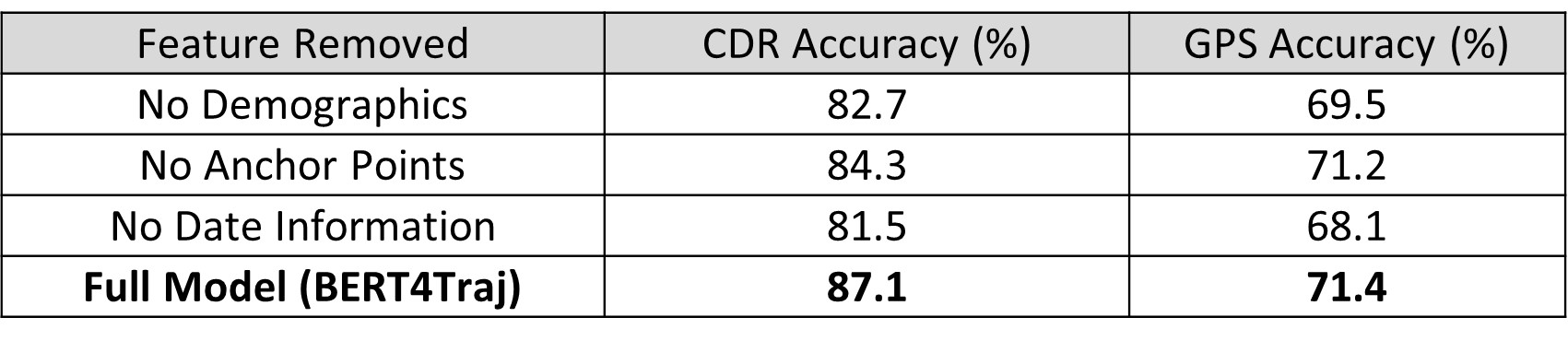}
    \label{tab:ablantion_study}
\end{table}

The findings show that removing any contextual feature leads to a decline in model performance. The most significant drop occurs when removing date information, reducing accuracy to 81.5\% in the CDR dataset and 68.1\% in the GPS dataset. This suggests that temporal context plays a crucial role in predicting missing locations. The removal of demographic data also results in a notable accuracy drop, indicating that user characteristics contribute valuable information for mobility prediction. Similarly, excluding anchor points reduces accuracy, highlighting their importance in modeling an individual’s movement behavior.

Overall, these results demonstrate that incorporating spatial, temporal, and demographic background information enhances the accuracy of BERT4Traj in reconstructing mobility trajectories.

\section{Conclusion}
This study introduced BERT4Traj, a Transformer-based model for reconstructing complete mobility trajectories from sparse location data. The model effectively captures spatial and temporal dynamic relationships, enabling more accurate trajectory reconstruction. Additionally, BERT4Traj enhances location prediction accuracy by incorporating multi-faceted contextual embeddings, including demographic, anchor point, and temporal features, enriching the representation of human mobility patterns. Moreover, BERT4Traj provides a scalable and adaptable framework for mobility datasets, making it applicable to public health, urban planning, and transportation analytics.

Despite its advantages, BERT4Traj has limitations. Its generalizability across different regions requires further validation, as mobility behaviors vary across geographic and socioeconomic contexts. Privacy concerns also emerge when reconstructing detailed trajectories, necessitating robust safeguards. Future research should explore multi-source mobility data integration, efficiency optimization, and privacy-preserving techniques. Addressing these challenges will enhance BERT4Traj’s reliability and applicability in human mobility research and decision-making.


\begin{thebibliography}{15}
\providecommand{\natexlab}[1]{#1}
\providecommand{\url}[1]{\texttt{#1}}
\expandafter\ifx\csname urlstyle\endcsname\relax
  \providecommand{\doi}[1]{doi: #1}\else
  \providecommand{\doi}{doi: \begingroup \urlstyle{rm}\Url}\fi

\bibitem[Barnett and Onnela(2020)]{barnett2020inferring}
Ian Barnett and Jukka-Pekka Onnela.
\newblock Inferring mobility measures from gps traces with missing data.
\newblock \emph{Biostatistics}, 21\penalty0 (2):\penalty0 e98--e112, 2020.

\bibitem[Belik et~al.(2011)Belik, Geisel, and Brockmann]{belik2011natural}
Vitaly Belik, Theo Geisel, and Dirk Brockmann.
\newblock Natural human mobility patterns and spatial spread of infectious diseases.
\newblock \emph{Physical Review X}, 1\penalty0 (1):\penalty0 011001, 2011.

\bibitem[Chen et~al.(2014)Chen, Bian, and Ma]{chen2014traces}
Cynthia Chen, Ling Bian, and Jingtao Ma.
\newblock From traces to trajectories: How well can we guess activity locations from mobile phone traces?
\newblock \emph{Transportation Research Part C: Emerging Technologies}, 46:\penalty0 326--337, 2014.

\bibitem[Chen et~al.(2019)Chen, Viana, Fiore, and Sarraute]{chen2019complete}
Guangshuo Chen, Aline~Carneiro Viana, Marco Fiore, and Carlos Sarraute.
\newblock Complete trajectory reconstruction from sparse mobile phone data.
\newblock \emph{EPJ Data Science}, 8\penalty0 (1):\penalty0 1--24, 2019.

\bibitem[Crivellari et~al.(2022)Crivellari, Resch, and Shi]{crivellari2022tracebert}
Alessandro Crivellari, Bernd Resch, and Yuhui Shi.
\newblock Tracebert—a feasibility study on reconstructing spatial--temporal gaps from incomplete motion trajectories via bert training process on discrete location sequences.
\newblock \emph{Sensors}, 22\penalty0 (4):\penalty0 1682, 2022.

\bibitem[Gambs et~al.(2012)Gambs, Killijian, and del Prado~Cortez]{gambs2012next}
S{\'e}bastien Gambs, Marc-Olivier Killijian, and Miguel~N{\'u}{\~n}ez del Prado~Cortez.
\newblock Next place prediction using mobility markov chains.
\newblock In \emph{Proceedings of the first workshop on measurement, privacy, and mobility}, pages 1--6, 2012.

\bibitem[Hoteit et~al.(2014)Hoteit, Secci, Sobolevsky, Ratti, and Pujolle]{hoteit2014estimating}
Sahar Hoteit, Stefano Secci, Stanislav Sobolevsky, Carlo Ratti, and Guy Pujolle.
\newblock Estimating human trajectories and hotspots through mobile phone data.
\newblock \emph{Computer Networks}, 64:\penalty0 296--307, 2014.

\bibitem[Huang et~al.(2015)Huang, Li, Liu, and Ban]{huang2015predicting}
Wei Huang, Songnian Li, Xintao Liu, and Yifang Ban.
\newblock Predicting human mobility with activity changes.
\newblock \emph{International Journal of Geographical Information Science}, 29\penalty0 (9):\penalty0 1569--1587, 2015.

\bibitem[Li et~al.(2019)Li, Gao, Lu, and Zhang]{li2019reconstruction}
Mingxiao Li, Song Gao, Feng Lu, and Hengcai Zhang.
\newblock Reconstruction of human movement trajectories from large-scale low-frequency mobile phone data.
\newblock \emph{Computers, Environment and Urban Systems}, 77:\penalty0 101346, 2019.

\bibitem[Liu et~al.(2018)Liu, Ma, Du, Pei, Yi, and Peng]{liu2018mapping}
Zhang Liu, Ting Ma, Yunyan Du, Tao Pei, Jiawei Yi, and Hui Peng.
\newblock Mapping hourly dynamics of urban population using trajectories reconstructed from mobile phone records.
\newblock \emph{Transactions in GIS}, 22\penalty0 (2):\penalty0 494--513, 2018.

\bibitem[Mai et~al.(2020)Mai, Janowicz, Yan, Zhu, Cai, and Lao]{mai2020multi}
Gengchen Mai, Krzysztof Janowicz, Bo~Yan, Rui Zhu, Ling Cai, and Ni~Lao.
\newblock Multi-scale representation learning for spatial feature distributions using grid cells.
\newblock \emph{arXiv preprint arXiv:2003.00824}, 2020.

\bibitem[Meloni et~al.(2011)Meloni, Perra, Arenas, G{\'o}mez, Moreno, and Vespignani]{meloni2011modeling}
Sandro Meloni, Nicola Perra, Alex Arenas, Sergio G{\'o}mez, Yamir Moreno, and Alessandro Vespignani.
\newblock Modeling human mobility responses to the large-scale spreading of infectious diseases.
\newblock \emph{Scientific reports}, 1\penalty0 (1):\penalty0 62, 2011.

\bibitem[Si et~al.(2023)Si, Yang, Xiang, Wang, Li, Zhang, Tu, and Chen]{si2023trajbert}
Junjun Si, Jin Yang, Yang Xiang, Hanqiu Wang, Li~Li, Rongqing Zhang, Bo~Tu, and Xiangqun Chen.
\newblock Trajbert: Bert-based trajectory recovery with spatial-temporal refinement for implicit sparse trajectories.
\newblock \emph{IEEE Transactions on Mobile Computing}, 2023.

\bibitem[Wang et~al.(2019)Wang, Wu, Lu, Zhao, and Feng]{wang2019deep}
Jingyuan Wang, Ning Wu, Xinxi Lu, Wayne~Xin Zhao, and Kai Feng.
\newblock Deep trajectory recovery with fine-grained calibration using kalman filter.
\newblock \emph{IEEE Transactions on Knowledge and Data Engineering}, 33\penalty0 (3):\penalty0 921--934, 2019.

\bibitem[Yu et~al.(2018)Yu, Russell, Mulholland, and Huang]{yu2018using}
Haofei Yu, Armistead Russell, James Mulholland, and Zhijiong Huang.
\newblock Using cell phone location to assess misclassification errors in air pollution exposure estimation.
\newblock \emph{Environmental pollution}, 233:\penalty0 261--266, 2018.

\end{thebibliography}
\end{document}